\documentclass[10pt,conference]{IEEEtran}
\IEEEoverridecommandlockouts
\usepackage{cite}
\usepackage{amsmath,amssymb,amsfonts}
\usepackage{algorithmic}
\usepackage{graphicx}
\usepackage{textcomp}
\usepackage{url}
\usepackage{xcolor}

\def\BibTeX{{\rm B\kern-.05em{\sc i\kern-.025em b}\kern-.08em
    T\kern-.1667em\lower.7ex\hbox{E}\kern-.125emX}}
\begin{document}
\bstctlcite{IEEEexample:BSTcontrol}

\title{Incremental Online Learning Algorithms Comparison for Gesture and Visual Smart Sensors}

\author{\IEEEauthorblockN{
Alessandro Avi}
\IEEEauthorblockA{\textit{University of Trento} \\
\textit{Department of Industrial Engineering}\\
Trento, Italy \\
alessandro.avi@studenti.unitn.it}
\and
\IEEEauthorblockN{
Andrea Albanese}
\IEEEauthorblockA{\textit{University of Trento} \\
\textit{Department of Industrial Engineering}\\
Trento, Italy \\
andrea.albanese@unitn.it}
\and
\IEEEauthorblockN{
Davide Brunelli}
\IEEEauthorblockA{\textit{University of Trento} \\
\textit{Department of Industrial Engineering}\\
Trento, Italy \\
davide.brunelli@unitn.it}
}

\maketitle

\begin{abstract}
Tiny machine learning (TinyML) in IoT systems exploits MCUs as edge devices for data processing. However, traditional TinyML methods can only perform inference, limited to static environments or classes. Real case scenarios usually work in dynamic environments, thus drifting the context where the original neural model is no more suitable.
For this reason, pre-trained models reduce accuracy and reliability during their lifetime because the data recorded slowly becomes obsolete or new patterns appear. Continual learning strategies maintain the model up to date, with runtime fine-tuning of the parameters. This paper compares four state-of-the-art algorithms in two real applications: i) gesture recognition based on accelerometer data and ii) image classification. 
Our results confirm these systems' reliability and the feasibility of deploying them in tiny-memory MCUs, with a drop in the accuracy of a few percentage points with respect to the original models for unconstrained computing platforms.
\end{abstract}

\begin{IEEEkeywords}
dataset, neural networks, online learning, tinyML
\end{IEEEkeywords}

\section{Introduction}
Today's Internet of Things (IoT) technology relies on the massive use of cloud computing resources to elaborate data generated by distributed objects and sensors. 
Improvements and more efficient applications have already been demonstrated by shifting the attention from cloud to edge, and distributing the computation along the IoT chain, including gateways and nodes~\cite{albanese2021automated, albanese2022low, brunelli2019energy}.
Using MCUs for intensive data elaboration can re-modulate part of the power consumption, which is crucial for these tiny devices. In fact, MCUs can optimize the data transmission and the data elaboration, which leads to the generation of more compact and meaningful information, and decreases the traffic of data on IoT networks. Moreover, edge computing improves the performance of real-time applications because the latency of communication is reduced, and feedback can be provided without cloud resources. \\
Among the different opportunities opened by edge computing, machine learning on tiny Embedded Systems, called TinyML, is gaining momentum. TinyML explores different types of models that can run on small, low-powered devices like microcontrollers for applications that require low latency, low power, and low bandwidth model inference.         
However, tiny devices feature limited memory and computing capability, which makes challenging the usage of ML models in edge devices. Thus, developing ML models of a few hundred Kbytes capable of keeping high accuracy while running MCU-enabled devices is still challenging. Another challenge is maintaining and upgrading deployed applications, which can be complex if devices are located in impractical positions.
Maintenance can be required for damage, upgrade, malfunction, and, more and more frequently, for neural network (NN) model update permitting the IoT device to evolve and correct its output. \\
Recently, continual learning (CL) systems have been introduced to update NN models on the MCU in real-time while the inference application runs. Current continual learning methods exploit continuous control over the error committed by the prediction to guarantee stable accuracy and reliable output. 
Such methods present some drawbacks, and the most important is called catastrophic forgetting~\cite{kirkpatrick2017overcoming}. This problem consists of a remarkable reduction of the classification accuracy of already learned classes while the system learns an additional class from new data.
Online or continuous learning applied to MCUs is a recent topic receiving growing attention, mainly because of the need to generate intelligent IoT systems that automatically self-upgrade, reducing the required maintenance. \\
This paper aims to test and compare the performance of state-of-the-art continual learning algorithms and propose changes to fit MCUs constraints. 
We did several tests using two different case studies, both targeting the classification of different types of data.  
The first is a typical gesture recognition application that classifies accelerometer data streams. We used an SMT32 Nucleo F401-RE equipped with an accelerometer shield as a test platform. This case study can be extended to other real-life applications such as industrial condition monitoring or anomaly detection~\cite{ren2021synergy, mostafavi2021novel}. 
The second case study compares different CL algorithms to classify instances from the MNIST dataset~\cite{deng2012mnist}. The tests are done on an OpenMV Cam H7 Plus, which is on an ARM Cortex M7. Even this experiment can be easily extended in real-life applications such as visual inspections of products in a manufacturing process. 
The performance comparison of CL algorithms is made starting on pre-trained models. 
All the algorithms use a small framework developed on the MCUs to modify and change the parameters of some layers (usually the last one) of the pre-trained model runtime. This makes the MCU an inference machine with training capabilities only on the selected layers. 
This paper presents the following contributions:
\begin{itemize}
    \item  Implementation of a framework for continual learning algorithms on STM32 MCUs and OpenMV Cam H7 Plus with micro-python interface. The framework extends the X-CUBE-AI expansion pack developed from STM for performing inference. CL algorithms use the error committed by the prediction to update the weights, save them in an array, and use the update rule defined by the selected CL algorithm. The algorithm's implementation is released as open-source software~\cite{release};
    \item Improvements of the most recent algorithms in the literature. The methods have been modified to apply backpropagation on the weights that depend on batches of incoming data;
    \item Comparison of state-of-the-art continuous learning algorithm performance in two case studies. 
\end{itemize}
The paper is organized as follows. Section~\ref{related_work} reviews the state-of-the-art studies on CL methods. Section~\ref{cl} presents the design of our framework. It briefly explains the used hardware, the algorithms, and the setup of the two case studies. Section~\ref{experimentalResults} describes the results of the experiments. Section~\ref{conclusion} concludes the paper.
\section{Related Works} 
\label{related_work}
TinyML has registered an impressive rate of scientific literature in the last few years.
So far, the main topics are the implementation of frameworks for optimal compression and the deployment of models for efficient inference. For example, MCUNet~\cite{lin2020mcunet} was one of the first to achieve high accuracy on off-the-shelf commercial microcontrollers, while other methods present deep compression with pruning~\cite{han2015deep} or SRAM-optimized approaches useful for the entire workflow in an ML model, including classification, porting, stitching and deployment~\cite{sudharsan2021sram}.
The work~\cite{ray2021review} proposes a complete review of the basic concepts of TinyML. The article contains information regarding some powerful and commonly used frameworks developed by research groups or corporations like Tensorflow Lite and X-CUBE-AI, the commonly used hardware and MCU, and some use case applications in which TinyML is particularly useful.  \\
In recent times, TinyML has expanded, and a relevant new branch related to Tiny online learning (TinyOL) has gained increasing attention. CL has already been explored for classic server and parallel architectures, but only in the last period focused on resource-constrained platforms.
As of now, there already exist some well performing strategies and frameworks like TinyTL~\cite{cai2020tinytl}, Progress \& Compress~\cite{schwarz2018progress}, TinyOL~\cite{ren2021tinyol}, and Train++~\cite{sudharsan2021train++}. \\
CL has already been successfully applied in a domain of interest for our study: image classification. For instance,~\cite{park2020convolutional} explore the usage of developmental memories for the damping of forgetting, and~\cite{disabato2020incremental} apply CL by using transfer learning and k-nearest neighbor. 
While applications of CL for pattern recognition on accelerometer sensors are still pretty new, this field has been explored only in some standard TinyML applications~\cite{zhou2021memory}. The studies mentioned above focus on creating memory and energy-efficient algorithms, and on backpropagation management and parameter manipulation.
The work presented in~\cite{grau2021device} applies CL on the edge with an exploration of federated learning, a method used for training distributed devices where ML models are trained locally and then merged into a global model. 
However, CL systems are not only related to training algorithms. For instance, Imbal-OL~\cite{sudharsan2021imbal} is a pre-processing technique that aims to remove unbalances of real-time data streams that are often present in real-life scenarios. The plugin can be added in between the input stream and the OL system to perform elaboration on the inputs and quickly adapt to changes while also preventing catastrophic forgetting. \\
CL strategies able to contrast catastrophic forgetting can be grouped into 4 categories~\cite{lesort2020continual}. The first one consists of architectural approaches that base their ability to overcome forgetting on the dynamics modification applied to the model's architecture. An example is the creation of dual memory models like in the CWR algorithm, where one weight matrix is used as short-term memory and another as long-term memory. The second category contains regularization approaches. These methods rely on adding some loss terms in the error computed from inference. The goal is to use these loss terms to redirect the backpropagation to maintain previously gathered knowledge while adapting to changes. Some of the most used and best-performing strategies of this kind are Elastic Weight Consolidation (EWC)~\cite{kirkpatrick2017overcoming}, Learn Without Forgetting (LWF)~\cite{li2017learning} and Synaptic Intelligence (SI)~\cite{zenke2017continual}. 
The last two strategy groups are the rehearsal and generative replay approaches. This study does not consider them because they require a high amount of memory and computations, unsuitable for TinyML applications. Rehearsal approaches are based on periodic refreshes of old data, which help the model avoid drifting from the original context. On the other hand, generative replay methods require using generative models to artificially generate past data to ensure that past knowledge is not forgotten.\\
CL systems can also be divided into two scenarios: multi-task (MT) and single-incremental-task (SIT). The first consists of learning tasks that are produced one by one in a controlled way without forgetting the previous ones. SIT is still an unexplored scenario and consists of neural networks that continuously expand their learning space that depends on the input data. SIT scenarios can lead to flexible systems that can add new classes to the NN design. However, the obtained systems are usually affected by catastrophic forgetting, which is a phenomenon that occurs when the knowledge related to past tasks is replaced with the newly-learned knowledge. 
Our study proposes the implementation of some regularization approaches aforementioned in SIT scenarios that can overcome catastrophic forgetting and are directly connected to the last layer of the NN model of interest~\cite{maltoni2019continuous}.
\section{System Design}  
\label{cl}
We started from the release of the TinyOL implementation presented in~\cite{ren2021tinyol}. The system can be attached to an already trained classification model to expand its learning capabilities. It permits both to learn never-seen classes and to fine-tune the parameters in the last layer in real-time. Differently from the system developed in~\cite{ren2021tinyol}, the proposed solution completely substitutes the last layer of the model in which the classification is performed, and it is attached to a truncated version of the pre-trained model, called the \emph{frozen model}. During the continual learning procedure, if the OL system detects a new class, it expands the dimension of the last layer, thus creating new weights and biases that are dedicated to the classification of the new class. In this study, because of the limited resources of the MCU, only the last layer is modified. The training is performed every time a new sample is received, with the goals of: i) fine-tuning the model to recognize and adapt to changes in old classes (known as New Instances settings, NI), ii) learning new patterns belonging to new classes never exposed to the system (known as New Classes setting, NC), iii) minimize the catastrophic forgetting~\cite{french1999catastrophic}.
We propose two examples characterized by New Instance and Classes (NIC) type data streams, meaning that the data contains both new classes and new instances of old classes. As described in~\cite{maltoni2019continuous}, this type of setting better represents a real-world application. \\ 
The OL system proposed in this paper consists of the frozen model and the OL layer. 
The main purpose of the frozen model is to perform the feature extraction of the input and later feed the elaborated data to the classification layer, which can change shape accordingly to the classes that it encounters. 
The second component is the OL layer, composed of a matrix that contains the weights and an array for the biases. The width and height of the weight matrix are directly dependent on the number of classes that the model can predict and the length of the last layer of the frozen model, while the shape of the bias array depends only on the number of known classes. Note that the initial values of the OL layer are not zeros or random values but just a copy of the original classification layer computed by Tensorflow during the frozen model training. \\
\subsection{Continuous Learning algorithms}
\begin{figure}[h]
      \centering 
        \includegraphics[width=0.9\columnwidth]{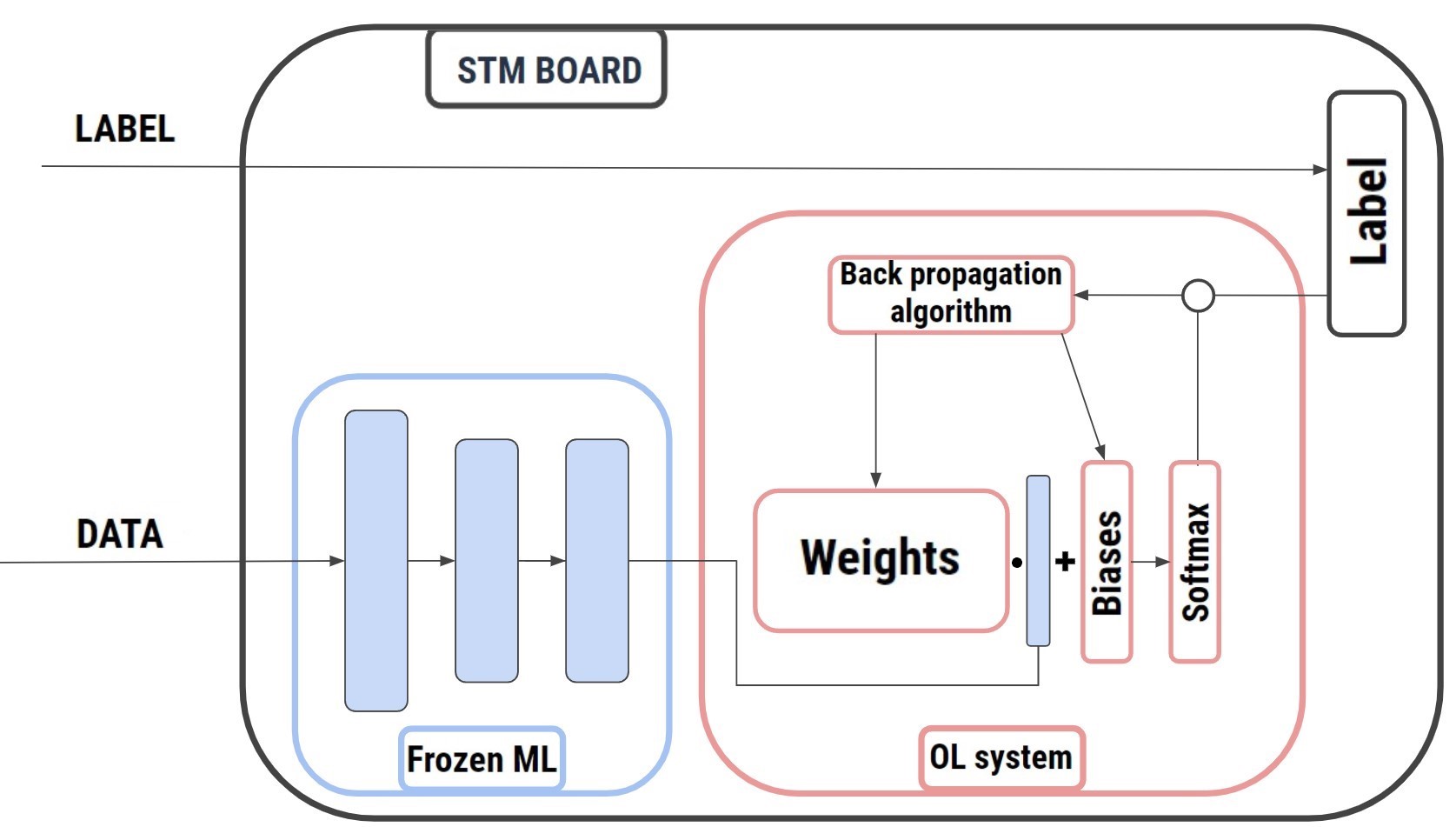}
        \caption{Basic block diagram of the OL system.}
        \label{fig:block_diagram}
        \vspace{-0.2cm}
    \end{figure}

The general structure of the proposed systems is similar across all the algorithms implemented. The training is performed in real-time in supervised mode, meaning that each time a data array and a ground truth label are received, a training step is performed on the OL layer. After this step, the data is discarded, saving memory in the device. \\
As shown in Figure~\ref{fig:block_diagram}, the data array and the label sent to the MCU are immediately elaborated by the frozen model. Before the OL layer starts the training, the label is checked. The goal is to control if the label received is already known by the system or not. If the label reveals to be an unknown class, the shape of the OL layer is enlarged, with a new row to the weight matrix, and a new cell to the bias array. This is possible because the OL layer is allocated at the boot in the RAM of the MCU, while the frozen model is usually stored in the flash memory of the device. After the feature extraction is performed by the frozen model, the output is fed to the OL layer. Here, the classification is performed with the formula $output=W \cdot frozen\_out + B$ and a softmax function applied to the output. The prediction is then compared with the ground truth label and, based on the error, the weights and biases are updated according to the specific algorithm rules. 
This paper focuses on the performance evaluation of some state-of-the-art algorithms designed for unconstrained platforms. Our contribution also includes the optimization of low-memory microcontrollers on the selected MCUs. The algorithms implemented are the standard NN training methods presented in~\cite{ren2021tinyol}. In details they are: i) a variation of TinyOL, namely TinyOL v2; ii) the  CWR algorithm initially proposed in~\cite{lomonaco2017core50} and later improved as CWR+ in~\cite{maltoni2019continuous}, and iii) the algorithm LWF shown in ~\cite{li2017learning, maltoni2019continuous}.
\subsubsection{TinyOL}
TinyOL exploits the standard approach used in NN training but applied in real-time contexts~\cite{ren2021tinyol}. Once the data passes through the frozen model, it arrives at the OL layer. Here, at first, the inference is performed, and later the training is done. The method uses gradient descent applied to the loss function, and a cross-entropy applied on the prediction and the ground truth label. 
%
%
Notice that the number of classes known from the OL system can change during runtime depending on the labels that are sent to the system; thus, the memory required for the OL layer can change unpredictably.\\
To overcome catastrophic forgetting, a variation of this method was implemented. We propose to exploit the information coming from a small batch and not just from the last sample read. In this case, the method requires the allocation of another weight matrix and bias array of the same size. The idea is to use the OL layer for inference, and each time a sample is received and elaborated by the model, its backpropagation is not applied directly to the OL layer but rather added in new containers called W and B (one for weights, the other for biases). When $k$ samples are received and elaborated by the system, the OL layer can then be updated using the average of the backpropagation data saved in W and B. Note that the user tunes the value $k$ that defines the size of the batch. The updates of the weights at each step become:
    \begin{align}
        W_{i,j} &= W_{i,j} + \alpha (y_i - t_i) \cdot x_j \label{weights2} \\
        B_i &= B_i + \alpha  (y_i - t_i) \label{biases2}
    \end{align}
At the end of each batch, the back propagation on the training weights is:
    \begin{align}
        w_{i,j} &= w_{i,j} - \frac{1}{batch\_size} \cdot W_{i,j} \label{weights3} \\ 
        b_i &= b_i - \frac{1}{batch\_size} \cdot B_i \label{biases3}
    \end{align}
Where $y_i$ is the prediction obtained from the OL layer, $t_i$ is the true label, $\alpha$ is the learning rate (tuned by the user), and $w_{i,j}$ and $b_i$ are the weights and the biases of the OL layer, respectively. 
The TinyOL method is a straightforward implementation of online training. It requires an amount of memory that depends only on the size of the weight and bias matrices, which depend on the number of classes known and the height of the last layer of the frozen model. The method allows the model to change its parameters each time new data arrives without any constraints. This is why the method cannot efficiently contrast the catastrophic forgetting.\\
TinyOL with mini-batches uses the same regularization approach but with double the memory. Unlike the TinyOL, this method tries to average the backpropagation over the last $k$ samples received. With a small value of $k$, the method is not contrasting the catastrophic forgetting but rather is trying to optimize the weight and bias variations to create a model that better predicts the outcome of batches. With a larger value of $k$, it can be possible to update weights, thus maintaining control over the forgetting. 
\subsubsection{TinyOL v2}
The method TinyOL v2 is a modified version of the original TinyOL. In this case, the method tackles the catastrophic forgetting of old classes, by not updating their weights and biases. This algorithm behaves as the TinyOL method except for the fact that the backpropagation is applied only if the weight or bias is related to a new class. Then, the backpropagation becomes: 
    \begin{align}
        w_{i,j} &= w_{i,j} - \alpha (y_i - t_i) \cdot x_j \label{OL2_weights}\\
        b_i &= b_i - \alpha (y_i - t_i) \label{OL2_biases}\\
        where \; i&=p,p+1..,n \; and \; j=0,1,..,m \nonumber
    \end{align}
%
In this case, the iterator $i$ starts from the value $p$, representing the first unknown class; thus, the update is performed only on the newest weights.
The TinyOL v2 is implemented with mini-batches to overcome catastrophic forgetting.\\ 
In conclusion, TinyOL v2 requires the same memory as the original TinyOL method but modifies only a portion of the OL layer to completely remove the catastrophic forgetting effect. 
However, this approach is not effective for fine-tuning the model when it tries to learn new patterns of the original classes. Additionally, the method does not optimize the classification layer for the prediction but separates its behavior into two parts where one stays updated while the other always remains the same, thus the weights of the last layer are not working together for the prediction.
The version TinyOL v2 with mini batches 
requires an amount of memory allocated slightly less than TinyOL mini batch because the matrix W and B 
are of shape $(n-p) \times m$ and $(n-p) \times 1$.
\subsubsection{LWF}
LWF is a regularization approach that overcomes catastrophic forgetting by balancing the backpropagation with new and old knowledge. This method requires the usage of two classification layers (each with a matrix of weights and an array of biases). The first one, called $tl$ (training layer), is continuously updated and performs the actual OL system inference. The second one, called $cl$ (copy layer), is the copy of the original frozen model classification layer. The idea of the algorithms is to perform the inference with both layers and then apply a loss function that depends on both outputs and the ground truth label. The gradient descent backpropagation is then applied to this loss function: 
    \begin{align}
        \mathcal{L}_{LWF} ( y_i, z_i, t_i) &=  (1-\lambda) \cdot \mathcal{L}_{cross}(y_i, t_i) + \lambda \cdot \mathcal{L}_{cross}(y_i, z_i) \label{LWF_loss}
    \end{align}
Where $y_i$ is the prediction array obtained from the layer $tl$, $z_i$ is the prediction array obtained from the layer $cl$, $t_i$ is the ground truth label, and $\lambda$ is the variable weight that defines which prediction has more relevance. In this algorithm, the loss function \eqref{LWF_loss} is composed of a weighted sum of the cross-entropy applied to two predictions, where the first part is related to the layer $tl$, which is always updated, and the second part is related to $cl$, which represents the original model. The role of $\lambda$ is extremely important because it defines which prediction can obtain more decisional power in the classification. The value of $\lambda$, as mentioned in~\cite{maltoni2019continuous}, cannot be maintained constant, but rather change together with the evolution of the training. With this, the LWF algorithm is a continuous balance between the continual learning and the original behavior. 
This application shows experimentally that evolving $\lambda$ with a function proportional to the number of predictions performed is a good solution. Thus, the update of the loss function weights is the following:
    \begin{align}
        \lambda = \frac{100}{100+ prediction \_ counter}  \label{LWF_lambda}
    \end{align}
For the sake of simplicity, this implementation follows the modification applied in~\cite{maltoni2019continuous}, where the loss function~\eqref{LWF_loss} is computed with both components being cross-entropy, 
and the other knowledge distillation loss. \\
A second version of the algorithm is proposed. The variation simply updates the values of $cl$ every $k$ training performed. This allows us to have a model that is a bit more flexible with respect to the LWF algorithm, where the two extremes are balanced (training layer and copy layer). The balancing is performed between the continually updated layer and a layer stopped in time (where its values are updated less frequently). In this case, the experimental lambda function found is defined by~\eqref{LWF_lambda2}. 
    \begin{align}
        \lambda = \frac{ batch \_ size}{prediction \_ counter}  \label{LWF_lambda2}
    \end{align}
Both the implemented LWF methods use the same amount of memory, which consists of two matrices of size $n \times m$ and two arrays of size $n \times 1$. 
This algorithm is more computationally expensive because it performs two predictions, which is one of its drawbacks. Nevertheless, the proposed method allows the model to overcome the problem of catastrophic forgetting. 
\subsubsection{CWR}
CWR is an approach that exploits the usage of two classification layers together with a weighted backpropagation method. 
The first one, called $tl$ (training layer), is updated every training step, and the second one, called $cl$ (consolidated layer), is updated once every batch.  During a training step, the inference is done only once with the $tl$,  and its weights and biases are updated with the standard TinyOL method. The breakthrough of the algorithm takes place at the end of a batch with the following rule, which is applied to both weights and biases of the layers: 
    \begin{align}
        cw_{i,j} &=  \frac{cw_{i,j} \cdot updates_{i} + tw_{i,j}}{updates_{i} + 1}  \label{CWR_cw}\\
        tw_{i,j} &=  cw_{i,j} \label{CWR_tw}
    \end{align}
Where $tw_{i,j}$ are the weights and biases of the training layer, $cw_{i,j}$ are the weights and biases of the consolidated layer, and $updates_{i}$ is an array that behaves as a counter of labels encountered. \\
The idea is to have a layer that changes slowly during the training and the other that is continuously updated. The layer $tl$ behaves as a short-term memory because it gets reset every time a batch ends, starting the training from a new point. On the other hand, the layer $cl$ behaves as the long-term memory because it never gets cleaned, and its weights and biases are updated using the $tl$. The update of the $cl$ layer is the particular backpropagation rule that allows the system to fuse slow memory and updated memory together. The method uses a sort of weighted average between weights where the weighting factor depends on the number of times a specific label appeared in the training batch. Note that the method requires the inference computation only from the $tl$ layer, while the other is actually never used. This is because while training the algorithm has no benefit in performing inferences with the $cl$ layer in training mode. In fact, an inference with this layer is done only if a prediction is requested, which happens only during testing.\\
CWR requires twice the memory of the TinyOL method because it allocates two matrices and two arrays of size $n \times m$ and $n \times 1$. 
The number of computations during training can double if a prediction is requested. Most times, the method executes only one inference per sample. When requested, the method also starts the $cl$ prediction, which is the most accurate of the two.
\subsection{Case Studies}
\begin{figure}[!h]
        \center
        \includegraphics[width=0.8\columnwidth]{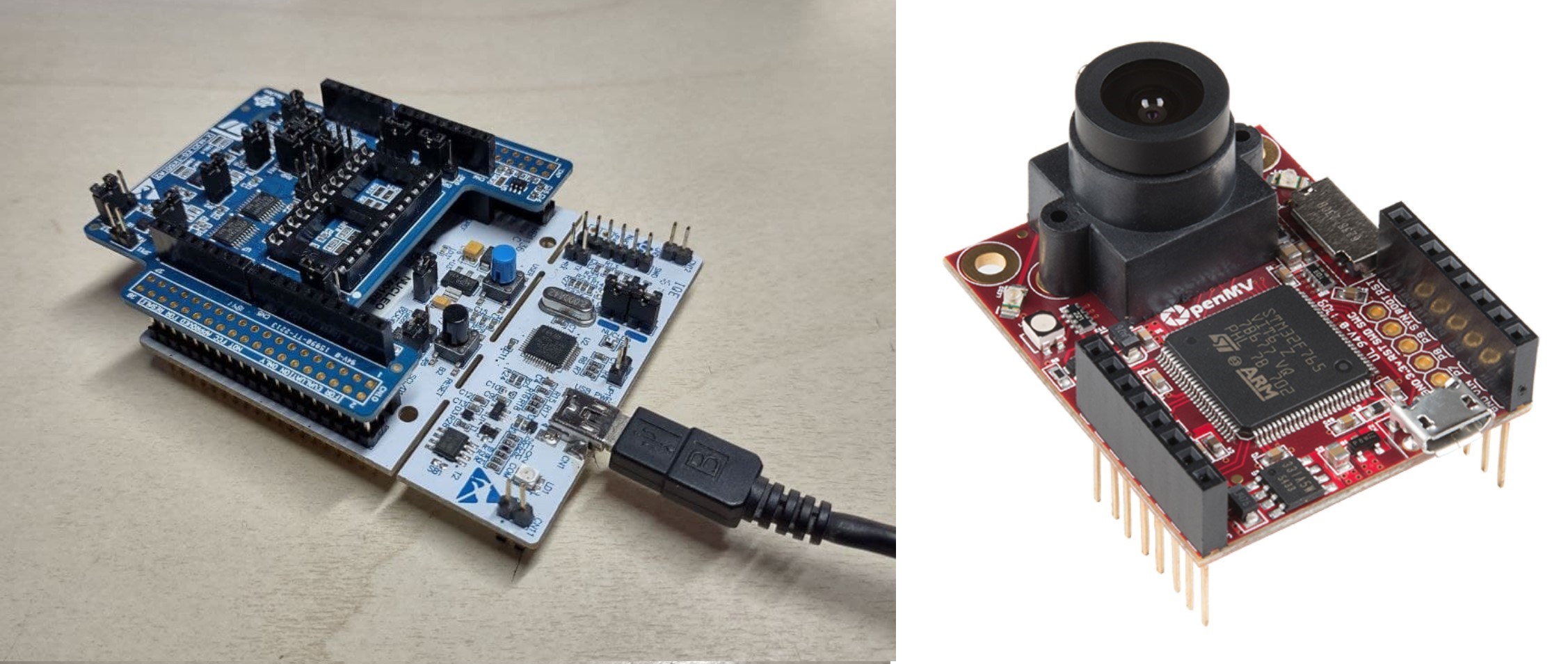}
        \caption{SMT32 F401RE paired with  acclelerometer shield on the left - OpenMV camera on the right.}
        \label{fig:hw}
    \end{figure}

To test and evaluate the algorithms described above, we used two applications. The first (experiment A) uses a Nucleo STM32 F401RE paired with a Nucleo shield IKS01A2 (Figure~\ref{fig:hw}), equipped with a 3D accelerometer sensor. This example aims to use an NN model on the data coming from the accelerometer to classify letters written in the air with the MCU mounted on the user's hand. The NN model is initially trained to recognize the vowels, later the OL system is attached to the model allowing it to learn three new letters B, R, and M. 
The second application (experiment B) uses an OpenMV cam (Figure~\ref{fig:hw}) as a visual smart sensor. It is based on the STM32H7 and permits the implementation of vision-based applications at low-power consumption. 
This application uses a CNN model that was initially trained to recognize the first 6 digits from the MNIST dataset~\cite{deng2012mnist}. The OL system is used to teach the model to recognize the remaining digits correctly.  
\subsubsection{Dataset Acquisition}
In experiment A, the dataset is acquired with the same hardware described above connected via USB to a laptop. The data is streamed in real-time via UART (USB cable) to the laptop, saving the data received in a text file. The acquisition of each letter lasts for 2 seconds, the MCU is set to work on a reading frequency of 100 Hz, making a single letter sample composed of 3 arrays (i.e., X, Y, Z accelerations) of 200 values each. 
All the letters in the dataset were recorded by the same user. The letters are always written in capitals following the same general path but with accentuated characteristics to make the dataset more similar to a NIC scenario (new classes and instances). 
The dataset is built on understanding 8 different letters, which are the original vowels A, E, I, O, U, and the additional consonants B, R, and M. Each vowel has a total of 560 samples, while the consonants have 760 samples each. The two groups' unbalanced samples are necessary because the vowels dataset is used in two pieces of training. The first portion is used for training the frozen model, which classifies only vowels. The remaining portion is added to the OL system training dataset, which also contains the letters B, R, and M. The final shapes of the two datasets are: 881 samples for the training of the frozen model and 4249 samples for the continual learning application. 
To make the data usable by the model, each sample is reshaped from a matrix of size $3 \times 200$ in an array of shapes $1 \times 600$. This is done simply by stacking each individual row horizontally. Once the datasets were correctly generated, they are also shuffled separately. \\
The application on the OpenMV camera relies on the usage of the well-known MNIST dataset~\cite{deng2012mnist}. The only pre-processing that was applied is the separation of the dataset into two groups called $low\_digits$ and $high\_digits$. The first group represents the digits from 0 to 5, which are used for the training of the CNN frozen model, while the remaining group represents the rest of the digits used in the OL training. The dataset used for the training of the frozen model contains a total of 36017 samples, while the dataset used for the OL training contains 5000 samples (500 for each digit). The reduced amount of samples in the dataset for the OL application is due to the time taken from the method itself. Anyway, the size selected is enough to guarantee fair results.
\subsubsection{Training and evaluation}
The model used for the accelerometer application is a NN classification model with low complexity and a low number of layers. The exact model structure is the following: input layer with 600 nodes, two hidden layers with 128 nodes and ReLu, output layer with 5 nodes, and Softmax. The frozen model is trained locally on a laptop using Tensorflow where the used optimizer is $Adam$, and the loss function is $categorical$ $cross$ $entropy$. The model is trained for 20 epochs with a batch size of 16. The testing shows a final accuracy of $96.83\%$. 
Another important step is the exportation of the trained model file, which is done in two versions. The first version is the simple exportation of the original model without modifications, while the second is the exportation of the truncated version (frozen model), where the last layer is removed. This is necessary because the OL system that we developed requires total control of the weights and biases of the last layer. Removing these values from the $model.h5$ and saving them in a text file in matrix form makes it possible to reload those in the code as matrices later. This allows the user to perform the standard inference from the $model.h5$ and later propagate the output through the last layer's weights, which are now accessible and editable.\\
The second application uses a CNN, which is more suited for image classification. The model structure is the following: two Conv2D 8 filters with Relu, MaxPooling 2x2, two Conv2D 8 filters with Relu, MaxPooling 2x2, Dropout 0.25, flatten, and Dense to 6 with Softmax. The frozen model is trained with Tensorflow with $Adam$ optimizer, $categorical$ $cross$ $entropy$ as loss function, 30 epochs, and a batch size of 64. The same exporting procedure presented above is performed. 
Note that, due to the limited memory on MCUs, pruning and quantization on the trained model is always suggested. In this case, the model structure does not require compression, but we decided to use it anyway to demonstrate the performance of these models.
\subsubsection{TinyOL Implementation} 
The OL layer is composed of one matrix for the weights and an array for the biases. These two containers are initialized at the beginning of the code. Their initial values are copied from a text file generated from the training step. In this way, it is possible to have a classification layer that starts exactly from where the Tensorflow training stopped. The matrix and array for experiment A have an initial shape of $5 \times 128$ and $1 \times 5$, respectively. 
The matrix and array for experiment B have an initial shape of $6 \times 512$ and $1 \times 6$, respectively. Note that in both applications, during training, each time a new class (letter or digit) is found, the OL system adds one row to the weights and one cell to the biases. This increases the allocated RAM memory, and it is important to ensure that the full capacity is not met. \\
The pseudo-test is done to test the performance of the CL algorithm of interest at running time. 
It is crucial to begin the test once a portion of the dataset has been processed to accurately represent the training method.
This type of testing better resembles a real-world application, where the OL system is deployed in an environment for an indefinite period of time, and its performance is checked online during runtime. In experiment A, the pseudo-testing starts when the training surpasses the $80\%$ of the dataset available. In experiment B, the testing starts when the sample 4000 is reached (on a total of 5000 samples).
%
\subsubsection{Evaluation metrics}
We measured the accuracy, training step time, and the maximum allocated RAM,  to evaluate the performance of the continual learning algorithms.
The accuracy indicates how much the model can predict incoming data and adapt to new data. Considering that this paper aims to prove the feasibility of deploying continual learning algorithms in MCUs, the accuracy can be enough to evaluate the models' goodness. The accuracy was computed with new data during testing; however, cross-validation can be considered for a more robust test.   
The training step time gives the time that the continual learning algorithm needs to compute a prediction and update its weights and biases. This metric assesses the algorithm responsiveness. Finally, the maximum allocated RAM permits a tradeoff between the usage of resources on the MCU and the achieved accuracy. These metrics allow a proper and efficient selection of the best algorithm that can run on embedded systems.
%
%
%
\section{Experimental Results}  
\label{experimentalResults}
\begin{figure}[ht]
      \centering 
        \includegraphics[width=1.0\columnwidth]{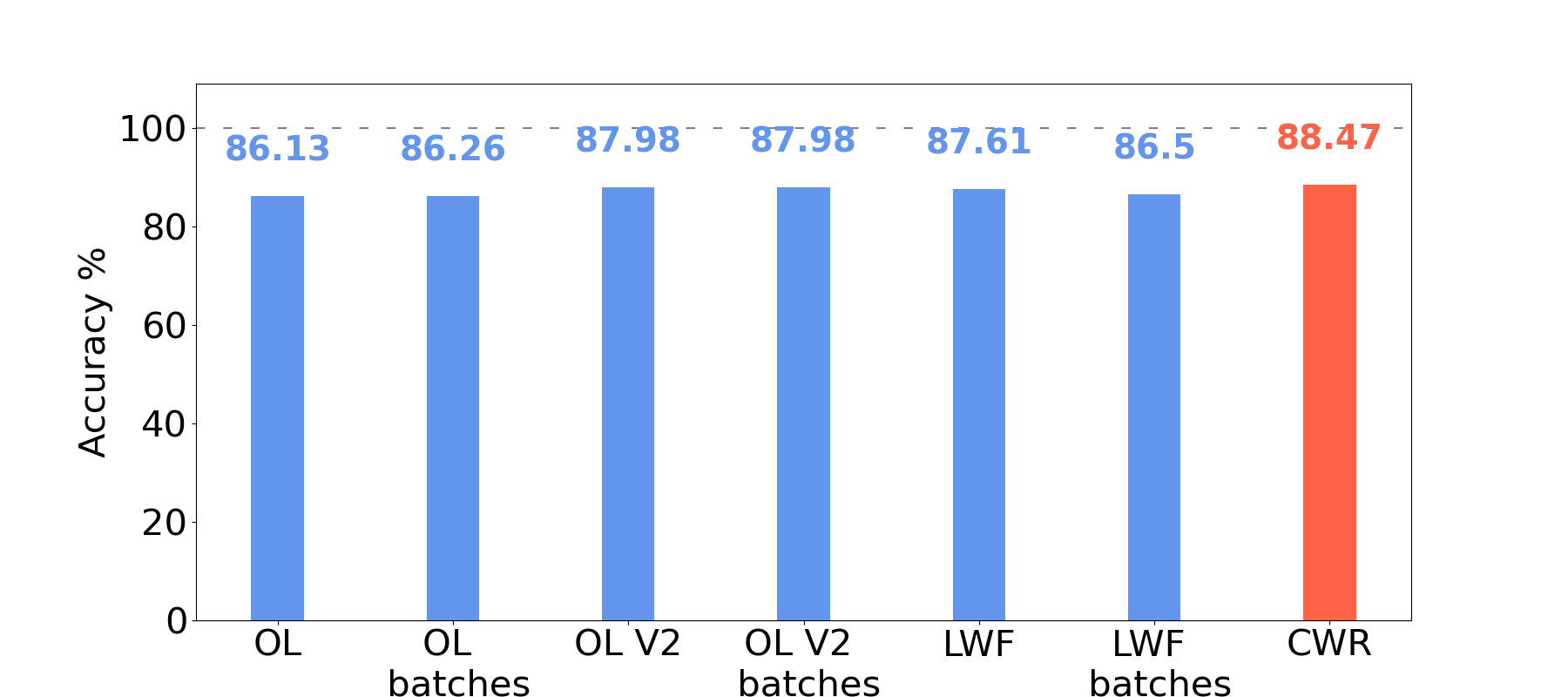}
        \caption{Accuracy of each strategy used - STM application.}
        \label{fig:letters_result}
    \end{figure}

Before benchmarking the proposed algorithms, we show a  comparison between results achieved on a laptop and those using the MCU. It is necessary to confirm that the performance is comparable and the algorithms are correctly optimized for the microcontrollers and capable of running online training reliably. 
After this, the results from the application on the OpenMV camera are shown, and the algorithm's accuracy is compared when new classes are added online.
The evolution of all the relevant parameters is stored from both devices and compared.
Parameter histories from both devices show the same behavior with a couple of exceptions that show a minor magnitude difference for just a few steps. From this test, it is possible to conclude that the MCU behaves in the same way as the laptop, proving that a device with such limited resources can be considered reliable and can be directly compared with the performance of more powerful devices.\\
Table~\ref{table:results} summarizes the performance of all algorithms from experiment A. Each test was performed with the same frozen model and dataset in the same order. The first row clearly shows that all the methods perform rather well, with the lowest accuracy being $86.13\%$, even though with a drop from the original Tensorflow training of $10.7\%$ due to the addition of the OL system. This means that a dataset of 4000 samples with approximately 500 samples for each class is enough for adequately training a model on three additional classes. Another important parameter is the inference time. The frozen model always behaves in the same way with a total inference time of $10.65\ ms$, no matter the strategy adopted. 
However, the inference time for the OL layer changes depending on the algorithm used. The slowest method is the LWF algorithm, which on average, requires more than double the time of all the other methods because it performs double the predictions for each training step, thus doubling the computations. All the other methods are very close, and the small differences are mainly due to different behaviors at the end of a batch. Another useful comparison is the RAM allocation. Table~\ref{table:results} confirms that the two lightweight methods are TinyOL and TinyOL v2, which allocate only one matrix and one array. The remaining methods use a similar amount of memory with a small variation of 100 bytes. Note that the Nucleo STM F401-RE MCU's total available RAM is 96 kB.  \\
\begin{table}[]
    \caption{Accuracy, average training step time, and memory allocated during training for all algorithms - STM application.}
    \label{table:results}
    \resizebox{\columnwidth}{!}{   
\begin{tabular}{l|c|c|c|c|c|c|c|}
\cline{2-8}
                                                                                                  & \textbf{TinyOL} & \textbf{\begin{tabular}[c]{@{}c@{}}TinyOL\\ batches\end{tabular}} & \textbf{\begin{tabular}[c]{@{}c@{}}TinyOL\\ V2\end{tabular}} & \textbf{\begin{tabular}[c]{@{}c@{}}TinyOL\\ V2 batches\end{tabular}} & \textbf{LWF} & \textbf{\begin{tabular}[c]{@{}c@{}}LWF\\ batches\end{tabular}} & \textbf{CWR} \\ \hline
\multicolumn{1}{|c|}{\textbf{\begin{tabular}[c]{@{}c@{}}Accuracy\\ (\%)\end{tabular}}}              & 86,13           & 86,26                                                             & 87,98                                                        & 87,98                                                                & 87,61        & 86,50                                                          & \textbf{88,47}       \\ \hline
\multicolumn{1}{|c|}{\textbf{\begin{tabular}[c]{@{}c@{}}Training step\\ time (ms)\end{tabular}}} & \textbf{0,99}            & 1,54                                                              & 1,03                                                         & 1,11                                                                 & 3,45         & 3,26                                                           & 2,11         \\ \hline
\multicolumn{1}{|l|}{\textbf{\begin{tabular}[c]{@{}l@{}}Max allocated\\ RAM (kB)\end{tabular}}}  & \textbf{26,1}            & 29,8                                                              & \textbf{26,1}                                                         & 29,8                                                                 & 29,9         & 29,9                                                           & 29,9         \\ \hline
\end{tabular}
    }
    \vspace{-0.6cm}
\end{table}
Better conclusions can be drawn from the plot in Figure~\ref{fig:letters_result} which displays the prediction accuracy for each method. 
The bar plot shows that all methods successfully perform OL training and maintain high accuracy. The method that shows the best overall accuracy is the CWR algorithm; however, the difference is very little. Thus it is taken into account also the inference time and the memory used. In general, all classes were correctly integrated into the OL model, and only some classification mistakes occurred in the letter R and B because of their similar shape.
Another important trend present in all algorithms is how the accuracy is spread quite uniformly among all classes. This confirms that all methods can perform continuous training and include new classes in their model structure. \\
The overall performance of the OL systems compared to the frozen model is reduced. This suggests that the accuracy of the original model is sacrificed to balance the system. This behavior can be tested even more by enlarging the dataset, which would help to understand if the model training has been early stopped. \\
Another important aspect concerns the \emph{catastrophic forgetting}. No method shows a strong effect against or in favor of it. Even the TinyOL method, which is considered to be the most vulnerable to the phenomenon, does not show poor behavior, meaning that the catastrophic forgetting in this specific application is not particularly severe. The phenomenon is probably dumped by the randomization of the dataset, which allows the models to refresh old data and learn new classes continuously. \\
Conclusions about the best performance consider all the characteristics such as accuracy, inference time, and memory used. The results shown here are obtained with learning rates tuned manually for each algorithm, and a common batch size of 16. \\

%
\vspace{-0.6cm}
    
%
%

\begin{figure}[h]
      \centering 
        \includegraphics[width=1.0\columnwidth]{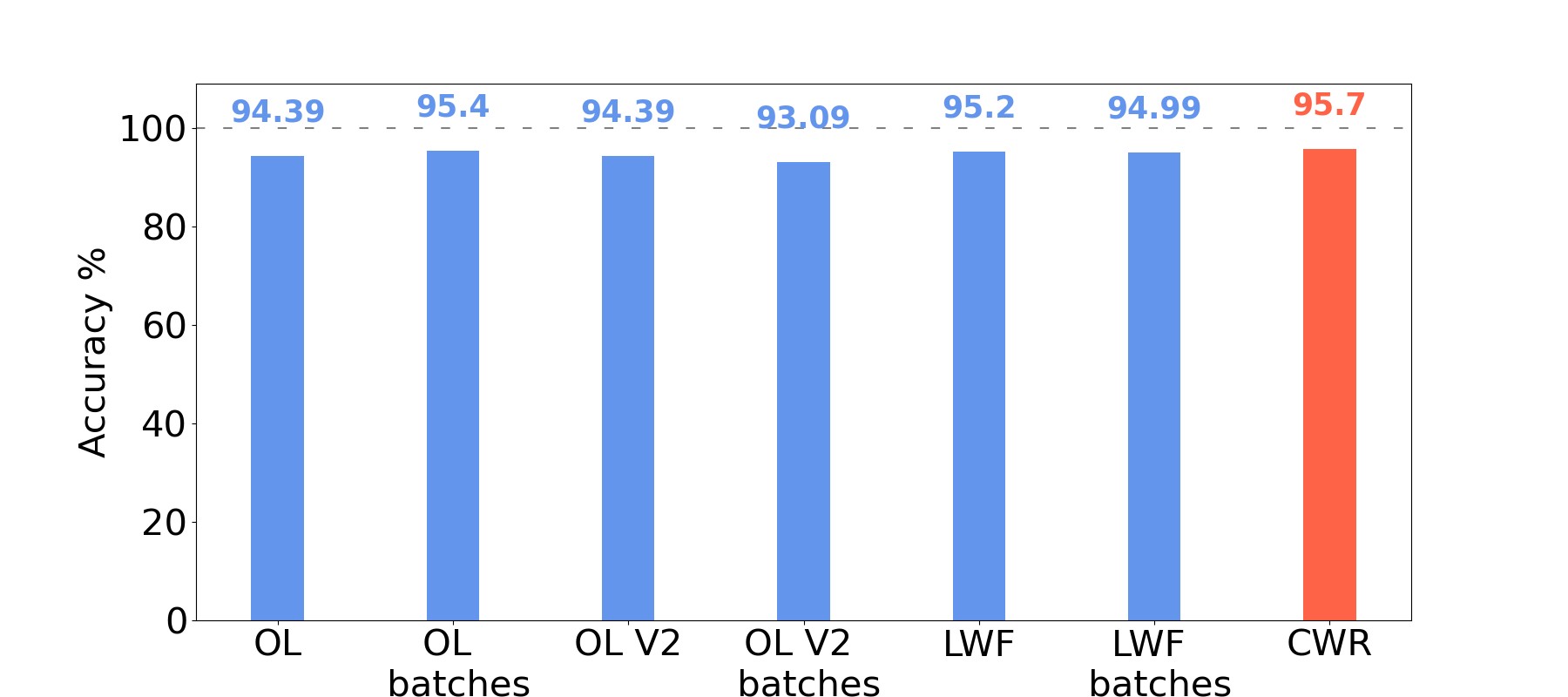}
        \caption{Accuracy of each strategy used - OpenMV application.}
        \label{fig:barplot_openmv}
        \vspace{-0.5cm}
    \end{figure}

\begin{figure}[ht]
      \centering 
        \includegraphics[width=1.0\columnwidth]{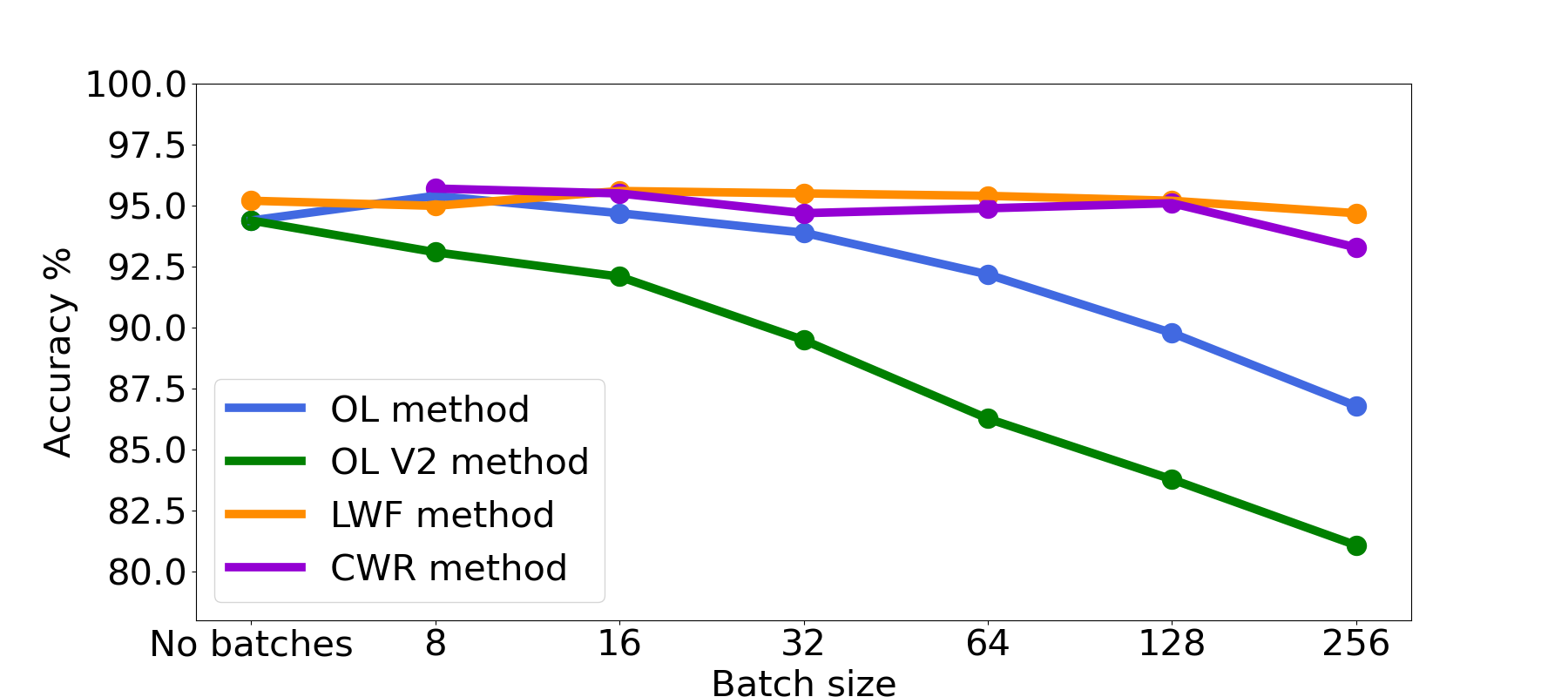}
        \caption{Accuracy of each class at variation of batch size - OpenMV application.}
        \label{fig:batch_result}
        \vspace{-0.6cm}
    \end{figure}

Concerning experiment B, the most relevant results are summarized in Table~\ref{table:results_openmv}. In this case, the test shows a higher accuracy compared to the previous experiment, most probably due to the high-quality dataset used. 
The inference performed by the frozen model stays pretty much constant for all methods and is about $15.88\ ms$. The increase of inference time for the OL layer in this application is given by the increased complexity of the frozen model. Instead, the inference time for the CL strategies is a bit different. All the methods that do not use batch updates are faster than their respective batch methods. 
Further conclusions can be drawn from the bar plots in Figure~\ref{fig:barplot_openmv}. 
In this case, the lowest accuracy is higher than the previous application, and in particular, the OL models have a much less drop in accuracy when compared to the Tensorflow training, from $99.35\%$ to $93.09\%$. Also, in this experiment, the presence of catastrophic forgetting is not severe. The most sensible method for the phenomenon (the OL method) does not show any problem due to the continuous refreshment of the old data.\\ 
Another important comparison can be done between the methods and their respective implementation with batches. 
\begin{table}[]
\caption{Accuracy and average training step time for all algorithms - OpenMV application.}
    \label{table:results_openmv}
    \resizebox{1.0\columnwidth}{!}{   
        \begin{tabular}{c|c|c|c|c|c|c|c|}
        \cline{2-8}
        \multicolumn{1}{l|}{} & \textbf{TinyOL} & \textbf{\begin{tabular}[c]{@{}c@{}}TinyOL\\ batches\end{tabular}} & \textbf{\begin{tabular}[c]{@{}c@{}}TinyOL\\ V2\end{tabular}} & \textbf{\begin{tabular}[c]{@{}c@{}}TinyOL\\ V2 batches\end{tabular}} & \textbf{LWF} & \textbf{\begin{tabular}[c]{@{}c@{}}LWF\\ batches\end{tabular}} & \textbf{CWR} \\ \hline
        \multicolumn{1}{|c|}{\textbf{\begin{tabular}[c]{@{}c@{}}Accuracy\\ (\%)\end{tabular}}} & 94,39 & 95,40 & 94,39 & 93,09 & 95,20 & 94,99 & \textbf{95,70} \\ \hline
        \multicolumn{1}{|c|}{\textbf{\begin{tabular}[c]{@{}c@{}}Training step\\ time (ms)\end{tabular}}} & 3,02 & 3,35 & \textbf{2,13} & 4,24 & 4,86 & 5,20 & 3,32 \\ 
        \hline
        \end{tabular}
    }
    \vspace{-0.6cm}
\end{table}
The accuracy difference in Table~\ref{table:results_openmv} is very small, but better conclusions can be drawn with Figure~\ref{fig:batch_result}, where the overall method accuracy is tested at the variation of the batch size. It is clear how the increase in batch size does not improve the model's accuracy. This suggests that the learning improvement (in particular for TinyOL and TinyOL v2) from just one sample is more significant than the averaged info obtained from a bigger group.
%

%
\section{Conclusions}
\label{conclusion}
This paper explores online learning on microcontrollers in two different scenarios concerning the analysis of accelerometer data and the classification of images. We adopted several state-of-the-art continual learning strategies with some improvements from our side. We showed that continual learning on small MCUs is a feasible solution for contrasting dynamic contexts, and the results obtained can generate self-sustainable and adaptable models. Our results demonstrate the capability of continual learning strategies to adapt to the extension of 4 new classes in the case of image classification or 3 classes in the case of gesture recognition. Models trained in these conditions can maintain high performance with an acceptable drop in accuracy 
of $10.7\%$ for the accelerometer example and $6.3\%$ for the MNIST digits recognition.\\
The obtained results demonstrate the potential of this type of technology, especially if applied on small and constrained devices. Continual learning applied in TinyML is a good alternative and an improvement to the standard approach of train-and-deploy because it permits flexible, self-adapting, and self-updating systems.

\section*{Acknowledgement}
This work was supported by the Italian Ministry for University and Research (MUR) under the program “Dipartimenti di Eccellenza (2018-2022)”.

\bibliography{acmart}

\begin{thebibliography}{10}
\providecommand{\url}[1]{#1}
\csname url@samestyle\endcsname
\providecommand{\newblock}{\relax}
\providecommand{\bibinfo}[2]{#2}
\providecommand{\BIBentrySTDinterwordspacing}{\spaceskip=0pt\relax}
\providecommand{\BIBentryALTinterwordstretchfactor}{4}
\providecommand{\BIBentryALTinterwordspacing}{\spaceskip=\fontdimen2\font plus
\BIBentryALTinterwordstretchfactor\fontdimen3\font minus
  \fontdimen4\font\relax}
\providecommand{\BIBforeignlanguage}[2]{{%
\expandafter\ifx\csname l@#1\endcsname\relax
\typeout{** WARNING: IEEEtran.bst: No hyphenation pattern has been}%
\typeout{** loaded for the language `#1'. Using the pattern for}%
\typeout{** the default language instead.}%
\else
\language=\csname l@#1\endcsname
\fi
#2}}
\providecommand{\BIBdecl}{\relax}
\BIBdecl
\renewcommand{\BIBentryALTinterwordstretchfactor}{4}

\bibitem{albanese2021automated}
A.~Albanese, M.~Nardello, and D.~Brunelli, ``Automated pest detection with dnn
  on the edge for precision agriculture,'' \emph{IEEE Journal on Emerging and
  Selected Topics in Circuits and Systems}, vol.~11, no.~3, pp. 458--467, 2021.

\bibitem{albanese2022low}
A.~Albanese, M.~Nardello, and D.~Brunelli, ``Low-power deep learning edge
  computing platform for resource constrained lightweight compact uavs,''
  \emph{Sustainable Computing: Informatics and Systems}, vol.~34, p. 100725,
  2022.

\bibitem{brunelli2019energy}
D.~Brunelli \emph{et~al.}, ``Energy neutral machine learning based iot device
  for pest detection in precision agriculture,'' \emph{IEEE Internet of Things
  Magazine}, vol.~2, no.~4, pp. 10--13, 2019.

\bibitem{kirkpatrick2017overcoming}
J.~Kirkpatrick \emph{et~al.}, ``Overcoming catastrophic forgetting in neural
  networks,'' \emph{Proceedings of the national academy of sciences}, vol. 114,
  no.~13, pp. 3521--3526, 2017.

\bibitem{ren2021synergy}
H.~Ren, D.~Anicic, and T.~Runkler, ``The synergy of complex event processing
  and tiny machine learning in industrial iot,'' \emph{arXiv
  preprint:2105.03371}, 2021.

\bibitem{mostafavi2021novel}
A.~Mostafavi and A.~Sadighi, ``A novel online machine learning approach for
  real-time condition monitoring of rotating machines,'' in \emph{2021 9th RSI
  International Conference on Robotics and Mechatronics (ICRoM)}.\hskip 1em
  plus 0.5em minus 0.4em\relax IEEE, 2021, pp. 267--273.

\bibitem{deng2012mnist}
L.~Deng, ``The mnist database of handwritten digit images for machine learning
  research [best of the web],'' \emph{IEEE Signal Processing Magazine},
  vol.~29, no.~6, pp. 141--142, 2012.

\bibitem{release}
\BIBentryALTinterwordspacing
A.~Avi, A.~Albanese, and D.~Brunelli. Incremental online learning for gesture
  and visual smart sensors. [Online]. Available:
  \url{https://gitlab.com/alba11/incremental-online-learning-for-gesture-and-visual-smart-sensors}
\BIBentrySTDinterwordspacing

\bibitem{lin2020mcunet}
J.~Lin \emph{et~al.}, ``Mcunet: Tiny deep learning on iot devices,''
  \emph{arXiv preprint:2007.10319}, 2020.

\bibitem{han2015deep}
S.~Han, H.~Mao, and W.~J. Dally, ``Deep compression: Compressing deep neural
  networks with pruning, trained quantization and huffman coding,'' \emph{arXiv
  preprint:1510.00149}, 2015.

\bibitem{sudharsan2021sram}
B.~Sudharsan \emph{et~al.}, ``An sram optimized approach for constant memory
  consumption and ultra-fast execution of ml classifiers on tinyml hardware,''
  in \emph{2021 IEEE International Conference on Services Computing
  (SCC)}.\hskip 1em plus 0.5em minus 0.4em\relax IEEE, 2021, pp. 319--328.

\bibitem{ray2021review}
P.~P. Ray, ``A review on tinyml: State-of-the-art and prospects,''
  \emph{Journal of King Saud University-Computer and Information Sciences},
  2021.

\bibitem{cai2020tinytl}
H.~Cai \emph{et~al.}, ``Tinytl: Reduce memory, not parameters for efficient
  on-device learning,'' \emph{arXiv preprint:2007.11622}, 2020.

\bibitem{schwarz2018progress}
J.~Schwarz \emph{et~al.}, ``Progress \& compress: A scalable framework for
  continual learning,'' in \emph{International Conference on Machine
  Learning}.\hskip 1em plus 0.5em minus 0.4em\relax PMLR, 2018, pp. 4528--4537.

\bibitem{ren2021tinyol}
H.~Ren, D.~Anicic, and T.~Runkler, ``Tinyol: Tinyml with online-learning on
  microcontrollers,'' \emph{arXiv preprint:2103.08295}, 2021.

\bibitem{sudharsan2021train++}
B.~Sudharsan \emph{et~al.}, ``Train++: An incremental ml model training
  algorithm to create self-learning iot devices,'' in \emph{2021 IEEE
  SmartWorld, Ubiquitous Intelligence \& Computing, Advanced \& Trusted
  Computing, Scalable Computing \& Communications, Internet of People and Smart
  City Innovation (SmartWorld/SCALCOM/UIC/ATC/IOP/SCI)}.\hskip 1em plus 0.5em
  minus 0.4em\relax IEEE, 2021, pp. 97--106.

\bibitem{park2020convolutional}
G.-M. Park, S.-M. Yoo, and J.-H. Kim, ``Convolutional neural network with
  developmental memory for continual learning,'' \emph{IEEE Transactions on
  Neural Networks and Learning Systems}, 2020.

\bibitem{disabato2020incremental}
S.~Disabato and M.~Roveri, ``Incremental on-device tiny machine learning,'' in
  \emph{Proceedings of the 2nd International Workshop on Challenges in
  Artificial Intelligence and Machine Learning for Internet of Things}, 2020,
  pp. 7--13.

\bibitem{zhou2021memory}
A.~Zhou, R.~Muller, and J.~Rabaey, ``Memory-efficient, limb position-aware hand
  gesture recognition using hyperdimensional computing,'' \emph{arXiv
  preprint:2103.05267}, 2021.

\bibitem{grau2021device}
M.~M. Grau, R.~P. Centelles, and F.~Freitag, ``On-device training of machine
  learning models on microcontrollers with a look at federated learning,'' in
  \emph{Proceedings of the Conference on Information Technology for Social
  Good}, 2021, pp. 198--203.

\bibitem{sudharsan2021imbal}
B.~Sudharsan, J.~G. Breslin, and M.~I. Ali, ``Imbal-ol: Online machine learning
  from imbalanced data streams in real-world iot,'' in \emph{2021 IEEE
  International Conference on Big Data (Big Data)}.\hskip 1em plus 0.5em minus
  0.4em\relax IEEE, 2021, pp. 4974--4978.

\bibitem{lesort2020continual}
T.~Lesort \emph{et~al.}, ``Continual learning for robotics: Definition,
  framework, learning strategies, opportunities and challenges,''
  \emph{Information fusion}, vol.~58, pp. 52--68, 2020.

\bibitem{li2017learning}
Z.~Li and D.~Hoiem, ``Learning without forgetting,'' \emph{IEEE transactions on
  pattern analysis and machine intelligence}, vol.~40, no.~12, pp. 2935--2947,
  2017.

\bibitem{zenke2017continual}
F.~Zenke, B.~Poole, and S.~Ganguli, ``Continual learning through synaptic
  intelligence,'' in \emph{International Conference on Machine Learning}.\hskip
  1em plus 0.5em minus 0.4em\relax PMLR, 2017, pp. 3987--3995.

\bibitem{maltoni2019continuous}
D.~Maltoni and V.~Lomonaco, ``Continuous learning in single-incremental-task
  scenarios,'' \emph{Neural Networks}, vol. 116, pp. 56--73, 2019.

\bibitem{french1999catastrophic}
R.~M. French, ``Catastrophic forgetting in connectionist networks,''
  \emph{Trends in cognitive sciences}, vol.~3, no.~4, pp. 128--135, 1999.

\bibitem{lomonaco2017core50}
V.~Lomonaco and D.~Maltoni, ``Core50: a new dataset and benchmark for
  continuous object recognition,'' in \emph{Conference on Robot
  Learning}.\hskip 1em plus 0.5em minus 0.4em\relax PMLR, 2017, pp. 17--26.

\end{thebibliography}
\bibliographystyle{IEEEtran}

\end{document}